%% file: arxiv.tex
\documentclass{article}
\pdfoutput=1
\usepackage{microtype}
\usepackage[dvipdfm]{graphicx}
\usepackage{subfigure}
\usepackage{booktabs} %

\usepackage{hyperref}

\usepackage{xcolor}
\definecolor{myblue}{HTML}{0060c0}
\definecolor{myred}{HTML}{c00060}
\definecolor{mygreen}{HTML}{60c000}

\hypersetup{
  linkcolor   = myblue,
  citecolor   = myred,
  urlcolor    = mygreen,
  colorlinks  = true,
  pagebackref = true,
}

\usepackage[utf8]{inputenc} %
\usepackage[T1]{fontenc}    %
\usepackage{url}            %
\usepackage{booktabs}       %
\usepackage{nicefrac}       %
\usepackage{microtype}      %

\usepackage{amsmath,amsfonts,amsthm,amssymb}
\usepackage[]{xcolor}
\usepackage[numbers]{natbib}
\usepackage{comment}
\usepackage{graphicx}

\DeclareRobustCommand*{\pf}{\overrightarrow{p}}
\DeclareRobustCommand*{\pb}{\overleftarrow{p}}

\newcommand{\xpf}{\overrightarrow{z}}
\newcommand{\xpb}{\overleftarrow{z}}
\newcommand{\Hf}{\overrightarrow{H}}
\newcommand{\Hb}{\overleftarrow{H}}
\newcommand{\yf}{\overrightarrow{y}}
\newcommand{\yb}{\overleftarrow{y}}

\makeatletter
\renewcommand\tableofcontents{%
    \@starttoc{toc}%
}
\makeatother
 \usepackage{authblk}

\begin{document}

\title{\mbox{Meet in the Middle: A New Pre-training Paradigm}}

\author[]{Anh Nguyen\thanks{Equal Contribution}} 
\author[]{Nikos Karampatziakis$^*$}
\author[]{Weizhu Chen}
\affil[]{Microsoft Azure AI}
\maketitle

\input{abstract}

\tableofcontents

\input{introduction}

\input{background}

\input{method}
\input{experiments}
\input{related_work}
\input{conclusion}

\bibliography{main}
\bibliographystyle{alpha}
\newpage
\appendix
\input{appendix}

\end{document}

%% file: abstract.tex
\begin{abstract}

Most language models (LMs) are trained and applied in an autoregressive 
left-to-right fashion, 
assuming that the next token only depends on the preceding ones. However, this assumption 
ignores the potential benefits of using the full sequence information during training, and the 
possibility of having context from both sides during inference. 
In this paper, we propose a new pre-training paradigm with techniques 
that jointly improve the training data efficiency and the 
capabilities of the LMs in the infilling task.  
The first is a training objective that aligns the predictions of a left-to-right LM with those of a right-to-left LM, trained on the same data but in reverse order. The second is a bidirectional inference procedure that enables both LMs to meet in the middle. We show the effectiveness of our pre-training paradigm with extensive experiments on both programming and natural language models, outperforming strong baselines. \footnote{Code and models available at \url{https://github.com/microsoft/Meet-in-the-Middle}}

\end{abstract}

%% file: introduction.tex
\section{Introduction} \label{sec:introduction}

Language models (LMs) are powerful tools for generating natural and
programming language, and have been widely used for various assisted
authoring tasks, such as text summarization, code completion, and
paraphrasing. In order to be usable in many different applications,
most LMs have to be able to generate the next token from the sequence 
of previous tokens. Given the importance of this operation, pre-training
has focused on optimizing the model's ability to predict the next token
given the previous tokens, as measured by perplexity. 
However, at pre-training time we have additional information that we are
not utilizing. In particular, when training the model to predict one token
we condition on the previous tokens (prefix) but completely ignore the subsequent 
tokens (suffix). While the suffix cannot be used as an input to 
the model, there are other ways to incorporate it into pre-training
which have not received attention in the literature. Our goal is
to utilize the pre-training data more efficiently
while preserving the autoregressive nature of the underlying LM. 

The approach we advocate involves additional modeling which at first blush 
may seem wasteful. After all, the main artifact produced during pre-training 
is an autoregressive left-to-right LM and the pre-training objective closely
matches how the LM is applied. Still, there are two reasons to consider alternative
training objectives. The first is about data efficiency.
The LM is trained by a cheap-to-obtain but rather sparse signal:
it produces a probability distribution over all possible choices 
for the next token 
yet it is only supervised by the actual next token in the training data.
What if during training we provided a denser form of supervision, where the probability distribution over next tokens is compared with another probability distribution? 

The second reason has to do with other 
related tasks. In particular, in many real-world
scenarios, the user may not want to generate text from scratch, but 
to rather infill or modify an existing sequence of tokens. For example, a
programmer may want to add a new argument to a function,
or a writer may want to insert a sentence or a phrase to improve the
coherence of a paragraph. 
In these cases, a left-to-right LM
cannot use the context from both sides of the insertion
position, and may produce suboptimal results.
The additional modeling
we do during training will also help us develop a state-of-the-art infilling technique.

In this work, we propose a unified pre-training and inference paradigm that we call 
``Meet in the Middle'' (MIM) to tackle both pre-training as 
well as infilling. MIM leverages two main ideas. The first idea 
is to introduce an additional language model that processes 
tokens right-to-left and use the two models to co-regularize 
each other. 
This allows each LM to benefit from the context provided by the other LM, which improves data efficiency and consistency.
Here the models ``meet in the middle'' metaphorically in the sense of adjusting 
their output probabilities to agree with the other side.
The second idea is a simple and effective inference procedure 
for infilling that takes advantage of all the artifacts 
produced during pre-training: both language models, as well 
as their tendency to agree. In this case, the two models will
be building the completion each from their own side until 
they literally ``meet in the middle''.
Our agreement regularizer has two important benefits: it regularizes the two language models and makes them more consistent, and it helps us stop the generation process early in the infilling task, by detecting when the two models converge to the same token.

In other words, to train MIM,
we use two decoding flows under a single shared decoder-only architecture~\citep{DBLP:conf/nips/BrownMRSKDNSSAA20},
\citep{DBLP:journals/corr/abs-2204-02311}.
The two LMs generate tokens in opposite directions. 
The forward direction predicts the next token 
given the prefix and the tokens it generates. 
The backward direction predicts the previous token 
given the suffix and the tokens it generates. 
We pre-train the two models jointly on a large corpus of text, using a combination of the standard language modeling loss and the agreement regularizer.
Once, pre-training is complete, the forward model 
is a drop-in replacement for existing autoregressive LMs. 
The backward model can either be discarded
or be used for related tasks such as infilling.

In our experiments, we aim to evaluate the effectiveness of MIM for pre-training LMs on different domains and tasks. We use public code and language data to pre-train LMs of different sizes and measure their performance in terms of perplexity and code completion tasks.
We compare MIM with FIM \cite{DBLP:journals/corr/abs-2207-14255} and other baselines, and show that MIM outperforms them in terms of both perplexity as well as 
task-specific evaluation metrics. We also conduct ablation studies to show the effectiveness of our main proposals during training and inference. To summarize, our main contributions are:

\begin{itemize}
    \item We introduce a new pre-training paradigm for LMs that uses the training data more efficiently by leveraging both the prefix and the suffix while still maintaining the autoregressive nature of LMs. We do this by training both a forward and a backward model and encourage them to agree.
    \item Propose a simple and efficient inference procedure for the infilling task, that takes advantage of context from both sides 
    and the tendency of the forward and backward models to agree.
    Our procedure can use parallelism more
    effectively than existing infilling procedures and on average achieves better quality and latency than the state of the art.
    \item Pre-train language models of different sizes on public code and language data using MIM, evaluate them both with human and programming languages, and show that MIM outperforms many baselines in terms of standard evaluation metrics. Finally, some models and code are made publicly available.
\end{itemize}

%% file: background.tex
\section{Preliminaries} \label{sec:pre}
We introduce some notation here we use throughout the paper. For a sequence
of tokens $x_1, x_2, \ldots, x_N$ we denote $x_{<i}$ the prefix
$x_1, x_2, \ldots x_{i-1}$. We use $x_{>i}$ for the suffix
$x_{i+1}, x_{i+2}, \ldots x_N$. The definitions for $x_{\leq i}$ and 
$x_{\geq i}$ are analogous. To reduce notation
clutter, we are suppressing all dependence of models on learnable 
parameters and when it is clear from context we even suppress the
inputs to the models. We will use arrows to distinguish the two 
models and their outputs. For example, $\pf$ is the forward model and 
$\pb$ is the backward model. Similarly, $\Hf$ will be a hidden 
representation from the forward model while $\Hb$ will be the 
corresponding representation from the backward model.

\subsection{The Infilling task}
In the infilling task we are given a sequence of tokens $x_1, x_2,\ldots, x_N$, an insertion position $i$ and a length $M$. 
The task is to generate a plausible (according to a LM) sequence of $M$ tokens $y_{1},\ldots,y_{M}$ to fill the gap between $x_{\leq i}$ and $x_{\geq i+1}$. 
In real world applications, $M$ is unknown. We consider it as an additional input  to avoid convoluted arguments about what constitutes a good infilling when $M$ is unknown. 
Given an autoregressive LM $p(x_t | x_1,\ldots, x_{t-1})$, a prefix $x_{\leq i}$, a suffix $x_{\geq i+1}$, and a length $M$, the task of finding the sequence $y_1,\ldots,y_M$, among all $M$ token sequences, that maximizes this probability requires time exponential in $M$. This is because a left-to-right LM cannot account for any disfluency that may occur between tokens $y_M$ and $x_{i+1}$. 

A simple technique for infilling that allows a LM to use context from both sides is called ``Fill in the Middle'' (FIM). \cite{DBLP:journals/corr/abs-2207-14255}. In FIM, 
the context for the LM is formed by concatenating the suffix and the prefix, in that order, to ensure coherence near the position where a completion is desired.
The advantages of the FIM approach are first, it can be applied to any pre-trained LM with very little modification, and second, it is computationally efficient, as it only requires one forward pass of the LM. However, the FIM approach also has some drawbacks namely the contexts are unnatural concatenations of tokens.
Furthermore, FIM cannot properly balance the influence of the prefix and the suffix, as the LM generation is typically biased towards the last few tokens of the context. 
With FIM we either have to move the suffix far from the point where the completion is requested, or (worse) move the prefix far from the completion.
Moreover, the FIM training procedure is itself ad-hoc as the documents are arbitrarily split into prefix, middle, and suffix, and the model is trained to predict the middle from the concatenation of suffix and prefix. The problem here is 
that during pre-training the model only sees one (or few) of the $O(N^2)$ possible (prefix, middle, suffix) splits in a document with $N$ tokens. 

\subsection{Bidirectional Language Modeling} \label{sec:bilm}

Bidirectional language modeling has been mainly used in the literature to train non-autoregressive LMs using training objectives such as Masked Language Modeling. Empirically, these non-autoregressive models seem to produce better representations than autoregressive LMs but have other disadvantages such as the difficulty to
perform in-context learning 
\cite{patel2022bidirectional}.

The first difference between our use of bidirectional modeling and the rest of the bidirectional modeling research is that
our model remains autoregressive. The future tokens are only used to regularize the model
and are not necessary for inference. The second difference is that we do not attempt to produce a 
single probability for every token. Instead there are always two probabilities, one computed from 
the past tokens (prefix) and one computed from the future tokens (suffix).

%% file: method.tex
\section{Meet in the Middle} \label{sec:method}

We now describe the details of our proposed solution first for pre-training 
and then for infilling. 
Intuitively, during pre-training we train two models that need to 
balance two goals: The first is that the models need to 
independently predict their next token well each using a different view
of the input (prefix vs. suffix).  The second  is that the probability 
distributions assigned to the next token from each model need to agree.
This gives each model a glimpse of its future and provides a more dense
supervision signal than simply the prediction of the next token.
Thus for pre-training we encourage the two models to ``meet in the middle'' 
in the sense of reaching a compromise between what
is predicted from the prefix and what is predicted from the suffix.

\subsection{Pre-training} \label{sec:pretraining}

We use two decoder-only language models that share all of their parameters, and we train both a forward model $\pf$
and a backward model $\pb$. The forward model $\pf$ is trained to predict next tokens in the
forward direction $x_1,x_2,\ldots$ and the backward model $\pb$ is trained to predict previous tokens in the backward direction $x_N, x_{N-1}, \ldots $. In other words,  $\pf$ is trained to maximize the 
likelihood of $x_t$ given $x_1,x_2,\ldots,x_{t-1}$ and $\pb$ is trained
to maximize the likelihood of $x_t$ given $x_{N},x_{N-1}, \ldots, x_{t+1}$.

To improve data efficiency during training we employ a natural co-regularization term 
that encourages $\pf$ and $\pb$ to \emph{agree}
on their predicted probability distribution over the vocabulary for each token. Many choices are possible here, such as different $f$-divergences or even more 
stringent measures of agreement such as the Euclidean distance between the contextualized representations of the same token according to the two models. We do not explore these choices here and leave them for future work. For our purposes, we used the simple total variation 
distance to capture the disagreements among the two models on the $i$-th token:
\[
D_{i,x}^{TV}(\pf || \pb) = \frac{1}{2}\sum_{z \in \mathcal{V}} \left| \pf(z | x_{<i}) - \pb(z | x_{>i}) \right|
\]
where $\mathcal{V}$ is the vocabulary. The agreement regularizer is the sum of these distances over all sentences and tokens. The benefit of this regularizer is two-fold. First, it can provide the models with a denser supervision signal which improves data efficiency and helps us train a better autoregressive LM. The reason we say it is a dense supervision signal is because the probabilities of all tokens from one side are compared with the probabilities of all tokens from the other side. In contrast the traditional language modeling objective only assesses the probability of the token that is actually observed. The second benefit of the agreement regularizer is that it encourages the models to agree on their predictions. As we will argue later, this degree of agreement affects the efficiency of our inference technique in the infilling task. 

To sum up, for a dataset $S$ of sequences, the full training loss is 
\begin{equation}
\sum_{x\in S} \sum_{i=1}^{|x|} 
-\log\left(\pf(x_i|x_{<i})\right) -\log\left(\pb(x_i|x_{>i})\right)
 + \beta D_{i,x}^{TV}(\pf||\pb).
\end{equation}
The hyperparameter $\beta$ is set to $0.1$ in our experiments (except for ablation studies where $\beta=0$). Once pre-training is done, we can directly use $\pf$ as a left-to-right autoregressive LM.

\subsection{Infilling} \label{sec:mim}

We now turn our attention to the infilling problem and describe how our inference procedure works.
Afterwards, we will discuss some optional modifications to the model architecture and training 
that could be used if we are only interested in infilling and do not need $\pf$ to be a 
drop-in replacement for existing autoregressive LMs.

\subsubsection{Inference}

At inference time, our goal is to have an efficient and low-latency generation procedure. At a conceptual level, a naive procedure could work as follows: First generate candidates from both models.
As mentioned earlier, $M$ the length of the desired infilling is unknown in many applications. So in practice we would generate from the two models until each meets a condition (e.g. an application specific token, such as the newline or the EOS token is generated). If $\pf$ generates $\yf_1, \yf_2, \ldots, \yf_F$ 
and $\pb$ generates $\yb_1, \yb_2, \ldots, \yb_B$ then we would need to find the best stitching 
$\yf_1, \ldots, \yf_i, \yb_{j}, \ldots \yb_B$ among all $(i,j)$ pairs $1\leq i \leq F$, $1\leq j \leq B$. However,
this can be time-consuming as we would need to examine and assign a score to all $F\times B$ possible stitchings.
\begin{figure*}
    \centering
    \includegraphics[width=0.9\textwidth]{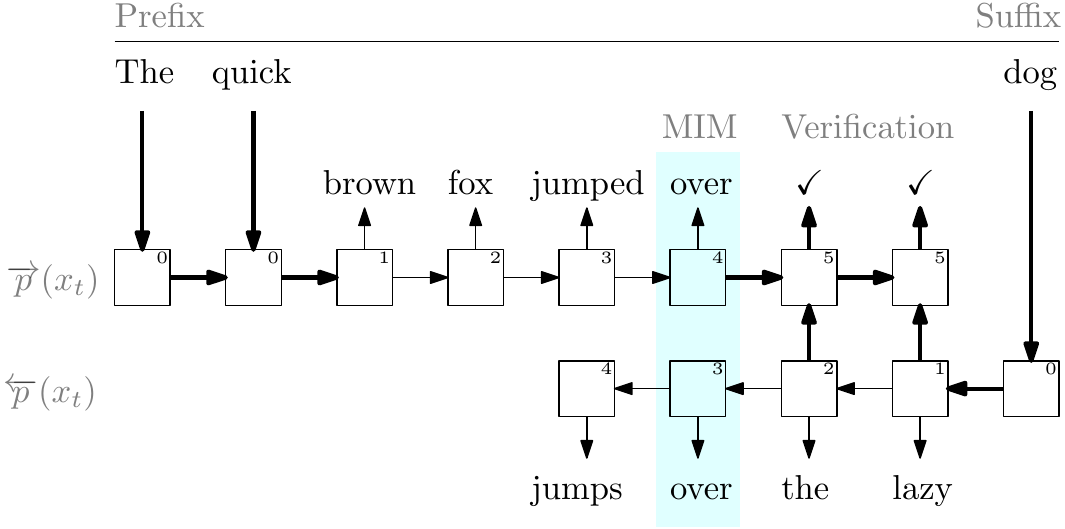}
    \caption{Inference procedure for infilling. Given the prefix ``The quick'' and the suffix ``dog'', the models $\pf$ and $\pb$ generate tokens until a candidate meet-in-the-middle token is detected (shaded area). We use a single token for illustration purposes, although the method can use more tokens. Given the candidate MIM token, the $\pf$ (respectively $\pb$) model can in parallel verify that the tokens generated by $\pb$ (resp. $\pf$) are acceptable completions (only the $\pf$ verification is shown to reduce clutter). The numbers in the top right of each box show the order of operations. Two boxes with the same number can be executed in parallel. Similarly, thick lines show that information (embeddings, tokens) can flow in parallel, while thin lines denote sequential steps.\label{fig:mim}}
\end{figure*}

Instead we propose a simplified procedure that interleaves generation and scoring and can terminate as quickly as a unidirectional LM. Moreover, with enough parallelism, it can even be faster. This approach is shown in Figure~\ref{fig:mim}. At a high level, the two models start building
a completion each from their own side and they try to literally meet in the middle.

The steps of our approach are as follows.
Initially, the prefix and suffix are consumed by
$\pf$ and $\pb$ respectively. Then $\pf$ and $\pb$ generate tokens synchronously one at a time. For each generated token from $\pf$ we check whether it is in the generated tokens from $\pb$ (or the first token of the suffix). Likewise, for each generated token from $\pb$ we check 
whether it is in the generated tokens from $\pf$ (or the last token of the prefix). If there is a match, we have a ``meet in the middle'' candidate position for joining the two generated sequences. 

In the most optimistic scenario, $\pf$ and $\pb$ produce the same sequence ($\pb$ produces it in reverse order). In that case we will detect that we can join the two generated sequences after each model has generated half of the tokens. Thus the two models have ``met in the middle''.  The importance of the agreement regularizer should now be more clear: if $\pf$ and $\pb$ produce completely different sequences, then our method is slower than FIM as it has no chance to terminate the generation early. But if the two models completely agree (and we have enough parallelism) it can even beat FIM in terms of generation latency as each model only needs to autoregressively generate half of the completion (while the other half is generated in parallel). As we will see in the experiments, our method achieves \emph{lower latency} than FIM which suggests that in most cases the two generated sequences meet somewhere near the middle.

Admittedly, the above procedure can suffer from false positives: just because one side generated ``the'' and the other side had generated ``the'' it does not necessarily mean that joining the two generated sequences at that position would produce a coherent infilling. To reduce false positives we use n-grams instead of a single token. In all of the infilling experiments, we use 4-gram matching. Finally, to be confident that the resulting sequence could have been generated by running $\pf$ or $\pb$ until the end, we run a parallel verification procedure which we describe next. 

Our parallel verification procedure is adapted from~\cite{DBLP:journals/corr/abs-2205-10350} where they used it 
in the context of certain tasks with special structure. 
Fortunately, our setup is the ideal place to apply their techniques. To make things more concrete, let's assume that 
the latest token generated from $\pf$ matches with a token in position $s$ generated from $\pb$, as is the case in Figure~\ref{fig:mim} for the token ``over''. Then we can use the tokens that were generated by $\pb$ before $s$ as inputs to $\pf$ \emph{in parallel}. In the context of Figure~\ref{fig:mim} we provide the tokens ``over'', ``the'', ``lazy'' as inputs to $\pf$ in parallel. 
If each output of $\pf$ matches the corresponding input, we have verified that $\pf$ would have autoregressively generated the tokens it copied from $\pb$'s output. Grounding the discussion back to Figure~\ref{fig:mim}, if $\pf$ generates (in parallel) ``the'' and ``lazy'' we have verified that $\pf$ would have autoregressively generated 
the same tokens. If there is
a partial match we can fast-forward the generation from $\pf$ to the point of the first disagreement. If there is no partial match
we can return to autoregressive generation from $\pf$ and $\pb$. So far, we have described verification as performing greedy top-1 sampling and checking whether the output from one position is the input we provided into the next position.
While this closely matches our parallel verification implementation, 
more relaxed acceptance criteria could be adopted such as the provided candidate token having a high enough probability in the previous output. 
If the generations from $\pf$ and $\pb$ terminate without meeting or passing verification, we return the sequence with higher probability according to the model that generated it.

To sum up, we generate by running both the forward and backward direction in parallel, and after each step check
whether there is a candidate meeting point where the two generated sequences are likely to be compatible. 
If so, we then apply a parallel verification procedure \citep{DBLP:conf/nips/SternSU18}, \citep{DBLP:journals/corr/abs-2205-10350} to the joined sequence and decide whether we should stop the generation process early. With this technique, ``Meet in the Middle'' (MIM) can produce high-quality outputs with \emph{better latency} than FIM.

\subsubsection{Optional Enhancements} \label{sec:optional}

To improve the infilling performance of our models, we can trade off their compatibility with autoregressive LMs and adopt a more powerful attention mechanism that allows bidirectional conditioning during generation. 
To do so, we switch the regular attention layer to a Synchronous Bidirectional Attention \cite{DBLP:journals/tacl/ZhouZZ19} layer which has recently shown promising results in Neural Machine Translation. This layer couples the forward and backward models together and allows the generation of one direction to condition on context and previously generated tokens from the opposite direction (in addition to its own context and previously generated tokens).

More concretely the Synchronous Bidirectional Attention \cite{DBLP:journals/tacl/ZhouZZ19} layer
modifies the regular attention layer activations in the following way: Let $\Hf$ be the output 
of an attention layer in $\pf$ and $\Hb$ the corresponding layer in $\pb$. Then
for a hyperparameter $\lambda$ we define the fused attention hidden representation as 
$$\Hf_{f} = \Hf +\lambda \Hb \textrm{ and } \Hb_{f}=\Hb + \lambda \Hf.$$ 
The models $\pf$ and $\pb$ then use $\Hf_{f}$ and $\Hb_{f}$ in place of $\Hf$ and $\Hb$. When 
$\lambda=0$, $\pf$ and $\pb$ are decoupled and reduce to classic autoregressive 
transformers.

With this modified architecture, care must be exercised during training. Recall 
that the LMs are trained with teacher forcing which means that during training
the prediction for each token is conditioned on the previous ground truth tokens 
rather than the previously generated tokens. Therefore, it is possible for the 
Synchronous Bidirectional Attention layer to leak information from 
the other direction during training. This could cause the LMs to simply 
learn to copy ground truth tokens from the other side. To prevent this
we make the training procedure somewhat closer to how inference is performed.
Similar to \citep{DBLP:journals/tacl/ZhouZZ19},
we employ a two stage process:
In the first stage we do not update the model parameters and simply use $\lambda=0$ to generate candidates $\xpf$, $\xpb$ from $\pf$ and $\pb$ respectively. We denote the candidates produced up to the $i-1$-th 
step of the first stage as $\xpf_{<i}$ and $\xpb_{<i}$. In the second stage we use 
the actual $\lambda>0$ and condition the probability of $x_i$ according to $\pf$ 
on both $x_{<i}$ and $\xpb_{<i}$. Similarly for $\pb$ we condition the probability of $x_{|x|-i}$ on both $x_{>|x|-i}$ and $\xpf_{<i}$, where $|x|$ is the length of sequence $x$.
Training is only done for the second stage. For the gradient computations, we treat 
$\xpf$ and $\xpb$ as constants even though they do depend on the current model parameters.

As we will show, the modifications suggested here further improve infilling metrics but come at the cost of incompatibility with existing autoregressive LMs. Therefore we consider them optional and 
application dependent.

%% file: experiments.tex
\section{Experiments} \label{sec:experiments}

This section presents the pre-training experiments, evaluation setup, main results and ablation studies for our models.

\subsection{Data and Models}
We first pre-train our models on a large and diverse corpus of public code with permissive licenses, which covers multiple programming languages. Python, Java, C++ are the dominant languages in our corpus, accounting for most of the pre-training data. After filtering and deduplication, our corpus contains about 300 billion tokens. Table~\ref{tab:corpus-stats} shows the detailed statistics of our pre-training data. We pre-train our models with a single pass over the data, processing about 300B tokens in total. This is about six times larger than the pre-training dataset used in the original Incoder model trained on 50B tokens including code and Stack Overflow data \citep{DBLP:journals/corr/abs-2204-05999}.

Apart from pre-training our model on public code datasets, we also pre-train our model on natural language, specifically the union of the following datasets: {\bf CC-News}, {\bf OpenWebText}, {\bf CC-Stories} and {\bf CC-100} with the following details. 
\begin{itemize}
    \item {\bf CC-News} contains 63 million English news articles crawled between September 2016 and February 2019 (76GB).
    \item {\bf OpenWebText} is an open source recreation of the WebText dataset used to train GPT-2 (38GB).
    \item {\bf CC-Stories} contains a subset of CommonCrawl data filtered to match the
story-like style of Winograd schemas (31GB).
    \item {\bf CC-100} is a dataset extracted from CommonCrawl snapshots between January 2018 and December 2018, filtered to match the style of Wikipedia (292GB) 
\end{itemize}

The first three of them are used to pre-train Roberta models \citep{DBLP:journals/corr/abs-1907-11692} and the rest is the English subset of CC-100 dataset, which in total contains 112B tokens. 

We adopt a decoder-only transformer language model \citep{DBLP:conf/nips/VaswaniSPUJGKP17, DBLP:conf/nips/BrownMRSKDNSSAA20} and train it to predict the next token in both directions, i.e., the same model and parameters are used for left-to-right and right-to-left prediction. To investigate the effect of model size, we pre-train models with three different capacities: 350M, 1.3B and 2.7B parameters. The hyperparameters and training setup for each model size are provided in the Appendix.

As baselines, following \citep{DBLP:journals/corr/abs-2207-14255}, we pre-train three FIM models with context-level FIM, apply transformations at the character level, use a FIM-rate of
$0.5$ and SPM+PSM joint training. Both the MIM and the FIM models are pre-trained using Megatron-LM framework \citep{DBLP:journals/corr/abs-1909-08053}. The model configurations for these variants as well as architectural details can be found in the Appendix.

\subsection{Benchmarks and metrics}
\subsubsection{Code generation and infilling}
We evaluate MIM in two different settings: Autoregressive left-to-right generation and infilling, and use different benchmarks and metrics for each setting.

To evaluate the autoregressive generation task, where the model needs to generate the code body given the function signature, docstring, and test cases, we use three widely used datasets of Python programming problems and one dataset containing multiple programming languages: {\bf HumanEval}~\citep{DBLP:journals/corr/abs-2107-03374}, which contains hand-crafted problems and solutions; {\bf MBPP}~\citep{DBLP:journals/corr/abs-2108-07732}, which consists of problems and solutions collected from crowd workers with a cleaned version that removes duplicates and errors; and {\bf APPS}~\citep{hendrycks2021measuring}, which has problems and solutions scraped from online coding platforms with varying difficulty levels. For the experiments with multiple programming languages, we use the {\bf HumanEval-X} dataset, a multilingual benchmark that contains 820 human-crafted coding problems in 5 programming languages, each of these problems is associated with tests and solutions. In this task, we only use the left-to-right model for inference, as there is no given suffix for the problems. 

For the infilling task, where the model needs to fill in the blank lines in an incomplete program, we use two recently proposed datasets that are designed for this scenario: {\bf HumanEval Infilling} \citep{DBLP:journals/corr/abs-2207-14255} and {\bf MBXP}  \citep{DBLP:journals/corr/abs-2210-14868}. This task mimics the common use cases of code completion tools like GitHub Copilot and document editors. Since these datasets provide suffixes and not all models can handle them properly, we compare MIM with Incoder, three variants of FIM, and code-davinci-002.

As for the metrics, we use the pass@k metrics \citep{DBLP:journals/corr/abs-2107-03374}, which measure the percentage of times that the generated code passes all the test cases within the top-k candidates. Specifically, we report pass@1, pass@10, and pass@100, and compute them using the unbiased estimator described in \cite{DBLP:journals/corr/abs-2107-03374}. For generation, we use top-p sampling with $p = 0.95$ and different temperatures: $0.2$ for pass@1 and $0.8$ for pass@10 and pass@100. Additionally, we report the single-line exact match (EM) metrics, which indicate the percentage of times that the completed lines exactly match the masked lines in the reference solution, as introduced in \citep{DBLP:journals/corr/abs-2204-05999} and adopted in \citep{DBLP:journals/corr/abs-2207-14255}.

\subsubsection{Language Modeling}
Apart from experiments on code generation and infilling, we also evaluate our models in the language modeling tasks to test the ability of our model to predict next token in a sequence measured by perplexity.

We evaluate and report perplexity in both {\bf in-domain} and {\bf out-of-domain} settings. For the {\bf in-domain} setting, we sample a held-out subset of the combined training data. For the {\bf out-of-domain} setting, we used the Pile dataset \citep{DBLP:journals/corr/abs-2101-00027}, a public language modeling dataset that combines data from various domains. We report the average perplexity across all subsets of the Pile dataset for our baselines and models.

\subsection{Main results}
\subsubsection{Code generation and infilling}
First, we present the results of our FIM model, pre-trained from scratch, in Table \ref{tab:main}, along with various baselines from the literature. Our FIM implementation significantly outperforms the Incoder models \citep{DBLP:journals/corr/abs-2207-14255}, on all metrics and datasets. For instance, our 2.7B model achieves 28.5\% pass@1 in HumanEval, while their 6.7B model only reaches 15.2\%. This impressive performance of our FIM model is attributed to several factors, such as the larger and better-filtered training data, and the implementation details we provide in the Appendix. Our FIM-2.7B model also surpasses other strong baselines, such as Codex 2.5B and CodeGen-Multi-6.1B, by a large margin. For example, FIM-2.7B attains 67.8\% in pass@100, compared to 44.9\% and 59.5\% of CodeGen-Multi-6.1B and Codex-2.5B, respectively. These results clearly establish that our FIM models are very competitive baselines. Therefore, in the following sections, we will focus on comparing MIM with our FIM baselines directly to highlight the benefits of "Meet-in-the-Middle".

Second, we evaluate MIM's autoregressive left-to-right generation in a prefix-only setting (no suffix). We compare FIM with MIM on both the HumanEval and MBPP benchmarks in Table \ref{tab:main}. Both of them have three different model sizes: 350M, 1.3B, and 2.7B. It is evident that MIM consistently outperforms FIM across all the metrics in both datasets. For example, the MIM-2.7B model boosts the HumanEval pass@1 to 30.7\%, surpassing the FIM-2.7B by 2.2\% (30.7\% vs. 28.5\%). The improvement on MBPP is similar, in terms of pass@1, 4.0\% (42.2\% vs. 38.2\%) on the 2.7B model and 0.9\% (26.8\% vs. 25.9\%) on the 1.3B model. We note that the improvements increase with the model size. This is evidence that the larger models benefit more from the agreement regularizer. We further evaluate this setting on the APPS dataset, which has three different difficulty levels. We report the results in Table \ref{tab:table2}. The trend and improvement are similar to the other two datasets, in which MIM consistently outperforms FIM across all the metrics and difficulty levels. In the multilingual setting, we compare MIM with FIM baselines in the HumanEval-X dataset. We observe consistent improvement across all metrics and all the programming languages that we evaluated on. Results are reported in Table \ref{tab:humaneval_x}. 

These results clearly demonstrate the advantage of MIM over FIM in the setting of left-to-right autoregressive generation.
The main claim in \citep{DBLP:journals/corr/abs-2207-14255} is that FIM does not harm the original left-to-right generative capability and can be learned for free. We argue that MIM is not only free, but also better in this setting. In other words, MIM pre-training which receives a more dense supervision from the agreement regularizer leads to a high quality left-to-right generative model and should become a new pre-training paradigm.

 Lastly, we evaluate MIM in the infilling setting, which is the setting that inspired us to design the MIM approach at the beginning. This is because in real-world applications, such as Copilot, we observe that the majority of usage is in developers jumping to the middle of a source file and then editing it with both the prefix and the suffix providing rich context. For this task, we reported results of both MIM and FIM baselines on the HumanEval Infilling benchmark and the MBXP Infilling benchmark in Table \ref{tab:main}. Comparing to the baselines, MIM consistently outperforms FIM in both datasets across all metrics and model sizes. Specifically, the MIM-2.7B model achieved pass@1 of $26.3\%$, an improvement of $3.5\%$ over FIM-2.7B model which achieved pass@1 of $22.8\%$. In terms of the exact match metric, an improvement of $6.1\%$ is also substantial ($57.8\%$ vs $51.7\%$). Similar to autoregressive left-to-right generation, we also notice improvement across three model sizes we consider.

\subsubsection{Language Modeling}

In this section, we compare FIM and MIM Models pre-trained on natural language datasets. We use  perplexity as our evaluation metric and look at two different language modeling settings, namely {\bf in-domain} and {\bf out-of-domain}. 

The perplexity results of both settings are summarized in Table \ref{tab:table3}. It is evident that across all experiments, MIM consistently outperform FIM baselines in terms of perplexity. The largest model MIM-2.7B has the best perplexity across all datasets in both settings. For example, MIM-2.7B model obtains a perplexity of 9.54 in OpenWebText dataset, a relative $19.9\%$ reduction in perplexity over FIM-2.7B model, which obtains a perplexity of 11.92. 

In the Pile dataset  \citep{DBLP:journals/corr/abs-2101-00027}, which represents the {\bf out-of-domain} setting, MIM-2.7B model also outperforms FIM-2.7B model by a relative reduction of $15.3\%$ in perplexity (9.24 vs 10.92), which further reinforces the advantages of MIM pre-training for natural languages.

\begin{table}[t]
    \centering
    \scalebox{0.6}{
        \begin{tabular}{lllllllllll}
        \toprule
        \multicolumn{1}{c}{{\textbf{Methods}}} & \multicolumn{3}{c}{{\textbf{HumanEval}}} & \multicolumn{3}{c}{{\textbf{MBPP}}} & \multicolumn{2}{c}{{\textbf{HE Infilling}}} & \multicolumn{1}{c}{{\textbf{MBXP}}} \\
        \cmidrule(lr){1-1}
        \cmidrule(lr){2-4}
        \cmidrule(lr){5-7}
        \cmidrule(lr){8-9}
        \cmidrule(lr){10-10}
        \multicolumn{1}{c}{$k$} &\multicolumn{1}{c}{$1$} & \multicolumn{1}{c}{$10$} & \multicolumn{1}{c}{$100$} & \multicolumn{1}{c}{$1$} &\multicolumn{1}{c}{$10$} & \multicolumn{1}{c}{$100$} & \multicolumn{1}{c}{$1$} &\multicolumn{1}{c}{\textsc{EM}} & \multicolumn{1}{c}{$1$} \\
        \midrule
         Incoder-1.3B & $8.9$ & $16.7$  & $25.6$  & $11.3$  & $26.8$  & $42.7$ & $8.6$ & $31$ & $9.2$ \\
         Incoder-6.7B & $15.2$ & $27.8$ & $47$ & $19.4$ & $46.5$ & $66.2$ & $14.5$ & $44.1$ & $20.8$\\
         CodeGen-Multi-6B & $18.2$ & $28.7$ & $44.9$ &-&-&-&-&-&-\\
         CodeGen-Multi-16B & $18.32$ & $32.07$ & $50.80$ &-&-&-&-&-&-\\
         Codex-2.5B & $21.36$ & $35.42$  & $59.5$ &-&-&-&-&-&- \\
         Codex-12B & $28.81$ & $46.81$  & $72.31$ & - & - & - &-&-&- \\
         code-davinci-001 & $39$ & $60.6$  & $84.1$ & $51.8$ & $72.8$ & $84.1$ &-&-&- \\
         code-davinci-002 & $47$ & $74.9$ & $94.1$ & $58.1$ & $76.7$ & $84.5$ & $51.3$ & $74$ & $57.6$\\
         \midrule
         PaLM-8B &  $3.6$ &-& $18.7$ & $5.0$ &-&-&-&-&-&\\
         PaLM-62B &  $15.9$ &-& $46.3$ & $21.4$ &-&-&-&-&-&\\
         PaLM-540B &  $26.2$ &-& $76.2$ & $36.8$ &-&-&-&-&-&\\
         \midrule
         LLaMA-7B &  $10.5$ &-& $36.5$ & $17.7$ &-&-&-&-&-&\\
         LLaMA-13B &  $15.8$ &-& $52.5$ & $22.0$ &-&-&-&-&-&\\
         LLaMA-33B &  $21.7$ &-& $70.7$ & $30.2$ &-&-&-&-&-&\\
         LLaMA-65B &  $23.7$ &-& $79.3$ & $37.7$ &-&-&-&-&-&\\
         \midrule
         FIM-350M & $12.8$ & $16.7$ & $27.8$ & $14.8$ & $30.2$ & $44.5$ & $11.8$ & $37.6$ & $14.3$\\
         FIM-1.3B & $20.8$ & $39.4$ & $51.7$ & $25.9$  & $45.6$ & $62.5$ & $15.7$ & $42.2$ & $22.1$\\
         FIM-2.7B & $28.5$ & $45.6$ & $67.8$ & $38.2$ & $61.2$ & $76.1$ & $22.8$ & $51.7$ & $30.4$\\
        \midrule
        MIM-350M & $13.7$ & $17.2$ & $28.5$ & $16.5$ & $33.7$ & $47.4$ & $14.6$ & $41.7$ & $16.4$ \\
         MIM-1.3B & $22.4$ & $41.7$ & $53.8$ & $26.8$ & $47.6$ & $65.1$  & $17.4$ & $47.6$ & $24.5$\\
         MIM-2.7B & \textbf{30.7} & \textbf{48.2} & \textbf{69.6} & \textbf{42.2} & \textbf{64.8} & \textbf{79.3} & \textbf{26.3} & \textbf{57.8} & \textbf{35.7}\\
        \midrule
        \end{tabular}
    }
    \caption{pass@k ($\%$) on the HumanEval, MBPP and HumanEval Infilling and MBXP benchmarks. Additionally, Exact Match (EM) metric \citep{DBLP:journals/corr/abs-2204-05999} is reported for HumanEval Infilling. FIM is the baseline `Fill in the Middle'' (FIM). MIM is our proposed bidirectional language modeling with Meet-in-the-middle. 
    For reference, we also report evaluation numbers of other models, namely Incoder-1.3B and Incoder-6.7B \citep{DBLP:journals/corr/abs-2204-05999}, Codex-12B \citep{DBLP:journals/corr/abs-2107-03374} ,code-davinci-001 and code-davinci-002. Note that, Codex-12B, davinci-001, PaLM \citep{DBLP:journals/corr/abs-2204-02311} and LLaMA \citep{DBLP:journals/corr/abs-2302-13971} were trained only with left-to-right autoregressive objective, thus, cannot perform infilling. 
    }
    \label{tab:main}
\end{table}

\begin{table}[t]
    \centering
    \scalebox{0.7}{
        \begin{tabular}{lllllllllllll}
        \toprule
        \multicolumn{1}{c}{{\textbf{Methods}}} & \multicolumn{3}{c}{{\textbf{C++}}} & \multicolumn{3}{c}{{\textbf{Java}}} & \multicolumn{3}{c}{{\textbf{Go}}}
        \\
        \cmidrule(lr){1-1}
        \cmidrule(lr){2-4}
        \cmidrule(lr){5-7}
        \cmidrule(lr){8-10}
        \multicolumn{1}{c}{$k$} &\multicolumn{1}{c}{$1$} & \multicolumn{1}{c}{$10$} & \multicolumn{1}{c}{$100$} & \multicolumn{1}{c}{$1$} &\multicolumn{1}{c}{$10$} & \multicolumn{1}{c}{$100$} & \multicolumn{1}{c}{$1$} &\multicolumn{1}{c}{$10$} & \multicolumn{1}{c}{$100$} \\
        \midrule
        Incoder-6B & $10.0$ & $20.0$ & $35.0$ & $9.0$ & $19.0$ & $40.0$ & $8.0$ & $14.0$ & $29.0$\\
         CodeGen-6B & $12.0$ & $20.0$ & $36.0$ & $15.0$ & $18.0$ & $40.0$ & $9.0$ & $22.0$ & $40.0$\\
        CodeGen-16B & $18.0$ & $30.0$ & $50.0$ & $15.0$ & $38.0$ & $60$ & $13.0$ & $25.0$ & $47.0$\\
        CodeGeeX-13B & $20.0$ & $31.0$ & $50.0$ & $16.0$ & $38.0$ & $58.0$ & $15.0$ & $25.0$ & $49.0$\\
       \midrule
         FIM-350M & $8.5$ & $18.3$ & $24.3$ & $11.3$ & $21.5$ & $27.6$ & $10.2$ & $16.8$ & $31.4$\\
         FIM-1.3B & $16.7$ & $31.5$ & $43.2$ & $15.4$ & $25.2$ & $32.6$ & $12.5$ & $23.7$ & $39.6$\\
         FIM-2.7B & $24.5$ & $38.6$ & $51.3$ & $18.3$ & $28.6$ & $38.7$ & $15.2$ & $30.4$ & $50.2$\\
        \midrule
        MIM-350M & $10.2$ & $19.6$ & $26.3$ & $11.1$ & $22.4$ & $28.2$ & $10.8$ & $17.5$ & $31.8$\\
         MIM-1.3B & $19.3$ & $36.5$ & $45.7$ & $17.6$ & $27.4$ & $34.7$  & $13.6$ & $25.1$ & $41.7$\\
         MIM-2.7B & \textbf{27.4} & \textbf{41.3} & \textbf{54.1} & \textbf{21.6} & \textbf{30.8} & \textbf{39.1} & \textbf{17.4} & \textbf{32.7} & \textbf{53.6} \\
         \midrule
        \end{tabular}
    }
    \caption{$pass@k$ ($\%$) results on the HumanEval-X benchmarks in the zero-shot settings for the baselines FIM and our MIM approach. Results of $k=1$, $10$ and $100$ are reported across all categories. 
    }
    \label{tab:humaneval_x}
\end{table}

\begin{table}[t]
    \centering
    \scalebox{0.7}{
        \begin{tabular}{llllllllll}
        \toprule
        \multicolumn{1}{c}{{\textbf{Methods}}} & \multicolumn{3}{c}{{\textbf{Introductory}}} & \multicolumn{3}{c}{{\textbf{Interview}}} & \multicolumn{3}{c}{{\textbf{Competition}}} \\
        \cmidrule(lr){1-1}
        \cmidrule(lr){2-4}
        \cmidrule(lr){5-7}
        \cmidrule(lr){8-10}
        \multicolumn{1}{c}{$k$} &\multicolumn{1}{c}{$1$} & \multicolumn{1}{c}{$10$} & \multicolumn{1}{c}{$100$} & \multicolumn{1}{c}{$1$} &\multicolumn{1}{c}{$10$} & \multicolumn{1}{c}{$100$} & \multicolumn{1}{c}{$1$} &\multicolumn{1}{c}{$10$} & \multicolumn{1}{c}{$100$} \\
        \midrule
         FIM-350M & $3.6$ & $7.8$ & $11.5$ & $0.0$ & $0.3$ & $1.2$ & $0.0$ & $0.04$ & $0.9$\\
         FIM-1.3B & $8.2$ & $11.6$ & $17.4$ & $0.12$ & $0.59$ & $1.9$ & $0.01$ & $0.07$ & $1.7$\\
         FIM-2.7B & $12.4$ & $15.7$ & $20.8$ & $0.27$ & $0.72$ & $2.4$ & $0.03$ & $0.095$ & $2.4$\\
        \midrule
        MIM-350M & $4.7$ & $9.2$ & $14.3$ & $0.2$ & $0.51$ & $2.3$ & $0.02$ & $0.06$ & $1.4$\\
         MIM-1.3B & $10.6$ & $14.2$ & $21.2$ & $0.36$ & $0.76$ & $3.6$  & $0.043$ & $0.09$ & $2.2$\\
         MIM-2.7B & \textbf{14.3} & \textbf{18.2} & \textbf{24.6} & \textbf{0.52} & \textbf{1.4} & \textbf{5.2} & \textbf{0.067} & \textbf{0.18} & \textbf{3.3} \\
         \midrule
        \end{tabular}
    }
    \caption{$pass@k$ ($\%$) results on the APPS benchmarks in the zero-shot settings for the baselines FIM and our MIM approach. Results of $k=1$, $10$ and $100$ are reported across all categories. 
    }
    \label{tab:table2}
\end{table}

\begin{table}[t]
    \centering
    \scalebox{0.65}{
        \begin{tabular}{llllll}
        \toprule
        \multicolumn{1}{c}{{\textbf{Methods}}} & \multicolumn{5}{c}
        {{\textbf{Datasets}}}
        \\
        \cmidrule(lr){1-1}
        \cmidrule(lr){2-6}
        \multicolumn{1}{c}{{\textsc{Models}}} &
        \multicolumn{1}{c}
        {{\textsc{CC-News}}} & \multicolumn{1}{c}{{\textsc{OpenWebText}}} & 
        \multicolumn{1}{c}{{\textsc{CC-Stories}}} &
        \multicolumn{1}{c}{{\textsc{CC-100}}} &
        \multicolumn{1}{c}{{\textsc{The Pile}}} \\
        \midrule
       FIM-350M & 21.16 & 17.91 & 20.89 & 14.23 & 15.14\\
         FIM-1.3B & 18.78 & 13.28 & 17.52 & 11.34 & 12.48\\
         FIM-2.7B & 13.45 & 11.92 & 13.43 & 9.43 & 10.92\\
         \midrule
         MIM-350M & 19.43 & 17.23 & 18.75 & 13.45 & 13.97\\
         MIM-1.3B & 16.04 & 12.23 & 14.63 & 11.16 & 10.79\\
         MIM-2.7B & \textbf{11.17} & \textbf{9.54} & \textbf{11.35} & \textbf{8.76} & \textbf{9.24}\\
        \midrule
      \end{tabular}
    }
    \caption{
    Perplexity results on all datasets in {\bf in-domain} setting including \textbf{CC-News}, \textbf{OpenWebText}, \textbf{CC-Stories}, and \textbf{CC-100} held-out datasets, and perplexity results in {\bf out-of-domain} setting - the Pile dataset \citep{DBLP:journals/corr/abs-2101-00027}
    }
    \label{tab:table3}
\end{table}

\subsection{Ablation Study}

\subsubsection{Effect of Optional Enhancements} \label{sec:ablation}
In this section, we perform an ablation study to qualitatively assess the effect of 
the optional enhancements we proposed in section~\ref{sec:optional} for the infilling task. 
We compare the purely autoregressive MIM model $(\lambda = 0)$ where the LMs are not allowed to observe generated tokens from the opposite side, and only utilize context from their own side during generation. We contrast this with using the Synchronous Bidirectional Attention layer with $\lambda=0.3$ that conditions on previously generated tokens from both sides. We used perplexity on the validation data to select the value $\lambda$. We reuse the same value of $\lambda$ during infilling to avoid any potential mismatch between training and inference. 

We conduct an experiment on the HumanEval Infilling benchmark \citep{DBLP:journals/corr/abs-2207-14255} and results are summarized in Table \ref{tab:ablation_bir}.
We notice that models that directly incorporate bidirectional context always outperform models that only utilize unidirectional context across all model sizes. As always, this is at the expense of the forward model no longer being a a drop-in replacement for standard autoregressive LMs.

\begin{table}[t]
    \centering
    \scalebox{0.9}{
        \begin{tabular}{lllll}
        \toprule
        \multicolumn{2}{c}{{\textbf{Methods}}} & \multicolumn{2}{c}{{\textbf{HE Infilling}}} & \multicolumn{1}{c}{{\textbf{MBXP}}} \\
        \cmidrule(lr){1-2}
        \cmidrule(lr){3-4}
        \cmidrule(lr){5-5}
        \multicolumn{1}{c}{\textsc{Models}} &
        \multicolumn{1}{c}{$\lambda$} &
        \multicolumn{1}{c}{pass@1} &
        \multicolumn{1}{c}{\textsc{EM}} & \multicolumn{1}{c}{pass@1} \\
        \midrule
         MIM-350M & $0.0$ & $12.5$ & $38.6$ & $14.1$ \\
         & $0.3$ & \textbf{14.6} & \textbf{41.7} & \textbf{16.4} \\
        \midrule
         MIM-1.3B & $0.0$ & $15.6$ & $45.2$ & $21.7$ \\
         & $0.3$ & \textbf{17.4} & \textbf{47.6} & \textbf{24.5}\\
        \midrule
         MIM-2.7B & $0.0$ & $24.7$ & $54.3$ & $32.4$ \\
         & $0.3$ & \textbf{26.3} & \textbf{57.8} & \textbf{35.7}\\
         \midrule
        \end{tabular}
    }
    \caption{
    pass@1 ($\%$) results of MIM without bidirectional context $(\lambda=0.0)$ and with bidirectional context $(\lambda=0.3)$ on the HumanEval Infilling and MBXP benchmarks. Exact match results on the HumanEval Infilling benchmark are also reported.
    }
    \label{tab:ablation_bir}
\end{table}

\subsubsection{Effect of Agreement Regularizer}
In this section, we perform an ablation study to qualitatively assess the importance of the token-level agreement regularizer of Section~\ref{sec:pretraining} during training. We show that encouraging agreement during training helps improve the infilling performance of our models in the HumanEval Infilling benchmark \citep{DBLP:journals/corr/abs-2207-14255}, as summarized in Table \ref{tab:ablation_approx}. 

Comparing with Table~\ref{tab:ablation_bir}, we see that models trained without token-level agreement regularization in general perform worse than models that do not utilize bidirectional context $(\lambda=0)$, which further emphasizes the importance of our 
agreement regularizer in making the predictions consistent between forward and backward directions. 

\begin{table}[t]
    \centering
    \scalebox{0.9}{
        \begin{tabular}{lllll}
        \toprule
        \multicolumn{2}{c}{{\textbf{Methods}}} & \multicolumn{2}{c}{{\textbf{HE Infilling}}} & \multicolumn{1}{c}{{\textbf{MBXP}}} \\
        \cmidrule(lr){1-2}
        \cmidrule(lr){3-4}
        \cmidrule(lr){5-5}
        \multicolumn{1}{c}{\textsc{Models}} &
        \multicolumn{1}{c}{Regularizer} &
        \multicolumn{1}{c}{pass@1}&
        \multicolumn{1}{c}{\textsc{EM}} & \multicolumn{1}{c}{pass@1}\\
        \midrule
         MIM-350M & no reg & $11.8$ & $35.9$ & $13.5$\\
         & + reg & \textbf{14.6} & \textbf{41.7} & \textbf{16.4} \\
        \midrule
         MIM-1.3B & no reg & $13.8$ & $42.7$ & $22.7$ \\
         & + reg & \textbf{17.4} & \textbf{47.6} & \textbf{24.5} \\
        \midrule
         MIM-2.7B & no reg & $23.2$ & $52.5$ & $30.6$\\
         & + reg & \textbf{26.3} & \textbf{57.8} & \textbf{35.7}\\
         \midrule
        \end{tabular}
    }
    \caption{
    pass@1 ($\%$) results of MIM with token-level agreement regularizer and without token-level agreement regularizer on the HumanEval Infilling and MBXP benchmark. Exact match results on the HumanEval Infilling benchmark are also reported.
    }
    \label{tab:ablation_approx}
\end{table}

\subsection{Efficiency of Inference}
\begin{figure*}[ht]
    \centering
    \vspace{2pt}
    \includegraphics[width=\linewidth]{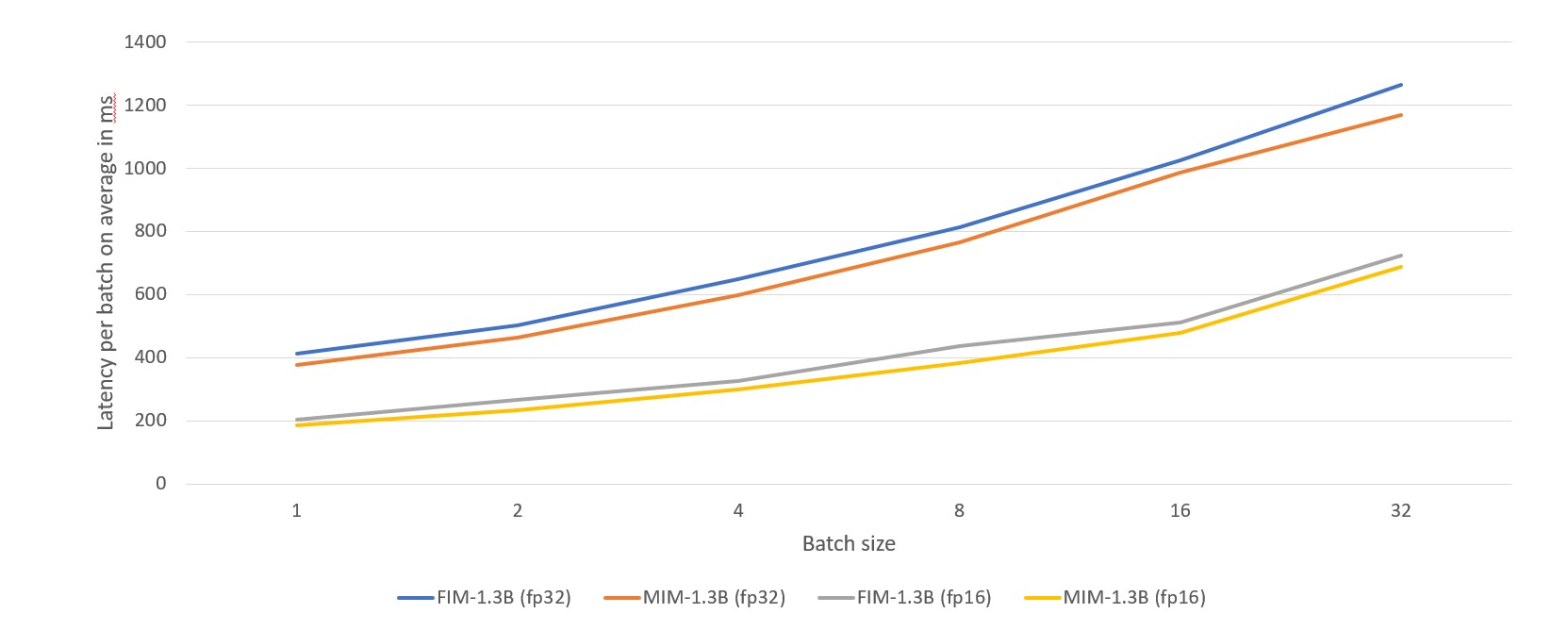}
    \caption{\small Inference latency of MIM and FIM baselines with batch implementation in A100 GPU with fp32 and fp16 precision.}
    \vspace{-3mm}
    \label{app:inference_latency}
\end{figure*}

In this section, we further look into the efficiency of our MIM inference procedure for infilling and compare to FIM in terms of inference latency with various batch sizes. Figure~\ref{app:inference_latency} shows the speedup of MIM in terms of inference latency
 over FIM baselines in both single and half precision format with the same model size of 1.3B average over all the examples in the HumanEval Infilling benchmark. In particular, the inference speed of MIM-1.3B is $4\%$ to $6\%$ faster compared to the inference speed of FIM-1.3B in single precision, and $3\%$ to $5\%$ faster if using half precision.

 We speculate that the speedup of MIM over FIM baselines during inference is attributed to several factors. Firstly, the generation of tokens in the left-to-right and right-to-left models are done in parallel. Furthermore, MIM inference procedure allows the generation from both sides to terminate early when there is an n-gram match and the sequence of tokens generated passes the verification. Our verification procedure based on \citep{DBLP:journals/corr/abs-2205-10350} is also very efficient as it can be parallelized over all the remaining time steps in the sequence.

%% file: related_work.tex
\section{Related work} \label{sec:related}

There is an extensive body of work on bidirectional language modeling. Early models such as BERT \citep{DBLP:conf/naacl/DevlinCLT19} masked tokens randomly, while T5 \citep{DBLP:journals/jmlr/RaffelSRLNMZLL20}, SpanBERT \citep{DBLP:journals/tacl/JoshiCLWZL20} masked spans of contiguous tokens and demonstrate improved performance. XLNET \citep{DBLP:conf/nips/YangDYCSL19}, on the other hand, utilizes bidirectional context during training by the permutation language modeling objective, which maximizes the likelihood over all factorization orderings of the training sequences. However, because these models typically focus on representation learning, in-context learning via prompting can be difficult \citep{patel2022bidirectional}.

Two works that train neural models using similar ideas are 
\cite{serdyuktwin} and \cite{zhang2019regularizing}. In the former
the authors train RNNs for acoustic modeling which needs to happen
in real-time. This constraint is very similar to our desire of 
having a forward model that can generate the next token from the 
previous ones. However, they propose to regularize the forward and
backward RNNs by requiring their representations to be close in 
Euclidean distance. We suspect that this constraint may be 
unnecessarily stringent, and trading it off against the 
LM's perplexity during training could hurt overall performance. 

On the other hand, \cite{zhang2019regularizing} only encourage
agreement in probability space. However, they are interested only in 
neural machine translation and only encourage agreement of 
probabilities at whole output sequence level. In contrast, we 
encourage agreement on every token. Another difference is that
they are using a sum of two KL divergence terms 
$KL(\pf||\pb)+KL(\pb||\pf)$ as the regularizer while we use
total variation distance. We suspect that KL is more 
stringent as it is unbounded and can hurt the overall model 
 when the model's perplexity needs to be traded off 
 against the regularizer during training.

Sharing all parameters between different factorizations of the sequence was first proposed in XLNET \citep{DBLP:conf/nips/YangDYCSL19} but has also been used with 
only forward and backward models in 
the context of image captioning \cite{zhou2022compact}.
That work motivated us to share 
all parameters between the forward and backward LMs.

Using LMs for infilling was first proposed in \cite{donahue2020enabling}
where the authors tackled the more
challenging setup with multiple places that need infilling.
\citep{DBLP:journals/corr/abs-2207-14255} applies ``Fill in the Middle'' (FIM) to the training data by randomly splitting each training instance into a tuple of (prefix, middle, suffix) and concatenate all these sections into a single example together with their sentinel tokens after tokenization. \citep{DBLP:journals/corr/abs-2207-14255} also contains a 
thorough section on research related to infilling. We refer the reader there for more details.

Furthermore, \cite{DBLP:journals/corr/abs-2201-07520} and   \citep{DBLP:journals/corr/abs-2204-05999} propose an extension of FIM, namely ``Causal Masked Language Modeling'' (CM3) and explore multi-region infilling problem. Similar to us, \citep{DBLP:journals/corr/abs-2207-14255} and \citep{DBLP:journals/corr/abs-2204-05999} leverage FIM to pre-train decoder-only language model on code data and evaluate their models on zero-shot code completion benchmark created from HumanEval dataset \citep{DBLP:journals/corr/abs-2107-03374}.

In \cite{sun2017bidirectional} the authors propose a bidirectional beam search procedure that also employs forward and backward LMs. However their focus is in improving accuracy at the expense of latency. Their procedure performs multiple passes over the completion, fixing tokens one-by-one and re-evaluating the 
probabilities from the LMs. Our approach focuses on achieving better infilling accuracy than FIM while also reducing latency.

%% file: conclusion.tex
\section{Conclusion} \label{sec:conclusion}

In this paper we addressed two challanges faced by large LMs:
Pre-training data efficiency and better handling of context 
for the task of infilling.
We proposed ``Meet in the Middle'', a method that uses 
both forward and backward LMs that share parameters 
and are trained to agree with each other in addition 
to predicting the next token. The resulting forward LM is 
a drop-in replacement for existing autoregressive LMs
while also achieving better quality over strong baselines.
Moreover, for the task of infilling, we proposed 
an inference procedure that employs both LMs and
can in certain cases reduce the inference latency by up to 50\%.
Though in our experiments the latency reduction was modest, compared to FIM, the reduction in perplexity and the improvements over FIM in both autoregressive and infilling settings were substantial.

\section*{Acknowledgments} \label{sec:ack}
We would like to thank Dejian Yang and Jian-Guang Lou at Microsoft Research Asia for helping us with processing training data. We also sincerely thank Daniel Fried and Armen Aghajanyan at Meta AI for helpful discussion regarding the training details of FIM baselines.

%% file: appendix.tex
\section{Appendix} \label{sec:appendix}

\subsection{Model training details} \label{app:training}
To evaluate the effectiveness of ``Meet in the Middle'' (MIM) pre-training compared to left-to-right autoregressive and ``Fill in the Middle'' (FIM) pre-training baselines, we adopt standard transformer-based autoregressive language models used in previous works \citep{DBLP:conf/nips/BrownMRSKDNSSAA20} for all the models we trained, varying the number of parameters (350M, 1.3B, 2.7B). Moreover, we replace the use of the Multi Head Attention \citep{DBLP:conf/nips/VaswaniSPUJGKP17} with the use of the Multi Query Attention proposed in \citep{DBLP:journals/corr/abs-1911-02150} in all the models we trained, allowing faster inference and reducing the memory requirements to store multiple key and values embeddings that are not shared between attention heads. 

For our bidirectional language models, we run the forward model and the backward  model in parallel within a single decoder-only architecture, leveraging bidirectional context explicitly during pre-training. 
We use the sentinel token $\langle l2r \rangle$ to specify that the generation comes from the forward model and sentinel token $\langle r2l \rangle$ to specify that generation comes from the backward model.

Regarding optimization, we use the Adam optimizer \citep{DBLP:journals/corr/KingmaB14} with $\beta_{1} = 0.9$, $\beta_{2} = 0.95$, $\epsilon = 10^{-8}$ and a global gradient norm clipping of $1.0$. We follow \citep{DBLP:conf/nips/BrownMRSKDNSSAA20} to decay learning rate to $10\%$ of its maximum value using cosine annealing with linear warm-up of $2\%$ of the total number of training steps.

For scaling the training of these models, we employ the open source Megatron-LM framework \citep{DBLP:journals/corr/abs-1909-08053} and partition the training across multiple GPUs along the batch dimension. All the training runs that we conducted use mixed precision training \citep{DBLP:conf/iclr/MicikeviciusNAD18} and FlashAttention \citep{DBLP:journals/corr/abs-2205-14135} to reduce memory requirements and increase training throughput. During pre-training of our models, we observed that MIM, FIM and autoregressive left-to-right pre-training have similar training wall-clock time, it is because the forward model and the backward model are executed in parallel in MIM pre-training. Our largest models of size 2.7B parameters are trained using 128 A100 GPU with 80GB memory each over 4 days, while the smaller models are trained using 64 A100 GPU with 80GB memory each over 3.5 days.
See Table \ref{tab:training_runs} for the details of all the training runs.

\subsection{Programming language dataset details} \label{app:data}

Table \ref{tab:corpus-stats} details the statistics of the datasets of different programming languages we use to pre-train our code language models in terms of number of tokens and dataset size. We perform some filtering and deduplication to obtain the final dataset. Our tokenizer is based on the Byte-Pair Encoding algorithm widely used in previous work \citep{DBLP:journals/corr/abs-2107-03374} to directly encode raw bytes with a vocabulary of size $100257$ tokens. We pre-tokenize the text using a special regex pattern that accounts for splitting on digit and newlines together with the default GPT-2 pre-tokenization \citep{DBLP:conf/nips/BrownMRSKDNSSAA20}.

\begin{table}[t]
    \centering
    \scalebox{1.0}{
        \begin{tabular}{lll}
        \toprule
        \multicolumn{1}{c}{{\textbf{Languages}}} & \multicolumn{1}{c}{{\textbf{Size (GB)}}} & \multicolumn{1}{c}{{\textbf{Tokens (B)}}} \\
        \cmidrule(lr){1-1}
        \cmidrule(lr){2-2}
        \cmidrule(lr){3-3}       
         C & $34.3$ & $12.3$ \\
         C++ & $215.6$ & $70.8$\\
         Python & $252.3$ & $75.5$\\
         Java & $178.5$ & $46.7$ \\
         JavaScript & $120.1$ & $39.3$\\
         TypeScript & $21.8$ & $8.6$ \\
         PHP & $30.7$ & $11$ \\
         Ruby & $26.8$ & $10.1$ \\
         C\# & $35.3$ & $12.6$ \\
         Others & $40.2$ & $13.3$ \\
         \midrule
         Total & $955.6$ & $300$ \\
        \midrule
        \end{tabular}
    }
    \caption{Approximate statistics of the programming language pre-training data}
    \label{tab:corpus-stats}
\end{table}

\begin{table}[t]
    \centering
    \scalebox{1.0}{
        \begin{tabular}{llll}
        \toprule
        \multicolumn{1}{c}{{\textbf{Hyper-parameters}}} & \multicolumn{1}{c}{{\textbf{350M}}} & \multicolumn{1}{c}{{\textbf{1.3B}}} & \multicolumn{1}{c}{{\textbf{2.7B}}}
        \\
        \cmidrule(lr){1-1}
        \cmidrule(lr){2-2}
        \cmidrule(lr){3-3}
        \cmidrule(lr){4-4}
        \\
        Number of layers & $24$ & $24$ & $32$ \\
        Number of heads & $16$ & $16$ & $32$ \\
        Dimension per head & $64$ & $128$ & $80$ \\
        Context length & $2048$ & $2048$ & $2048$\\
        Batch size & $786$k & $1$M & $1$M \\
        Weight decay & $0.1$ & $0.1$ & $0.1$\\
        Learning rate & $0.0003$ & $0.0002$ & $0.0002$ \\
        Warmup steps & $7$k & $5$k & $5$k\\
        Total steps & $382$k & $286$k & $286$k\\
        \midrule
        \end{tabular}
    }
    \caption{Details of each training run for all of our model specifications. 
    }
    \label{tab:training_runs}
\end{table}